\documentclass[10pt,twocolumn,letterpaper]{article}

\usepackage[pagenumbers]{cvpr} 

\usepackage{makecell}

\usepackage{mathrsfs}
\usepackage{lineno}
\usepackage{booktabs}
\usepackage{arydshln}
\usepackage{bbm}
\usepackage{threeparttable}
\usepackage{color,graphicx}
\usepackage{amsmath}
\usepackage{mathrsfs}
\usepackage{amsthm}
\usepackage[ruled]{algorithm2e}
\usepackage{bm}
\usepackage{multirow}
\usepackage{amssymb}
\usepackage{bm}
\usepackage{multirow}
\usepackage{arydshln}
\usepackage{amssymb}
\usepackage{nicematrix,booktabs,caption}
\usepackage[ruled]{algorithm2e}

\definecolor{cvprblue}{rgb}{0.21,0.49,0.74}
\usepackage[pagebackref,breaklinks,colorlinks,allcolors=cvprblue]{hyperref}

\def\paperID{2134} 
\def\confName{CVPR}
\def\confYear{2026}

\title{Endless World: Real-Time 3D-Aware Long Video Generation}

\author{Ke Zhang$^1$\quad
Yiqun Mei$^2$\quad
Jiacong Xu$^1$\quad
Vishal M. Patel$^1$\\
$^1$ Johns Hopkins University\quad$^2$ Adobe Research\\
\texttt{\{kzhang99,jxu155,vpatel36\}@jhu.edu
yiqun@adobe.com} \\
}
\begin{document}
\twocolumn[{%
\renewcommand\twocolumn[1][]{#1}%
\maketitle
\includegraphics[width=\textwidth]{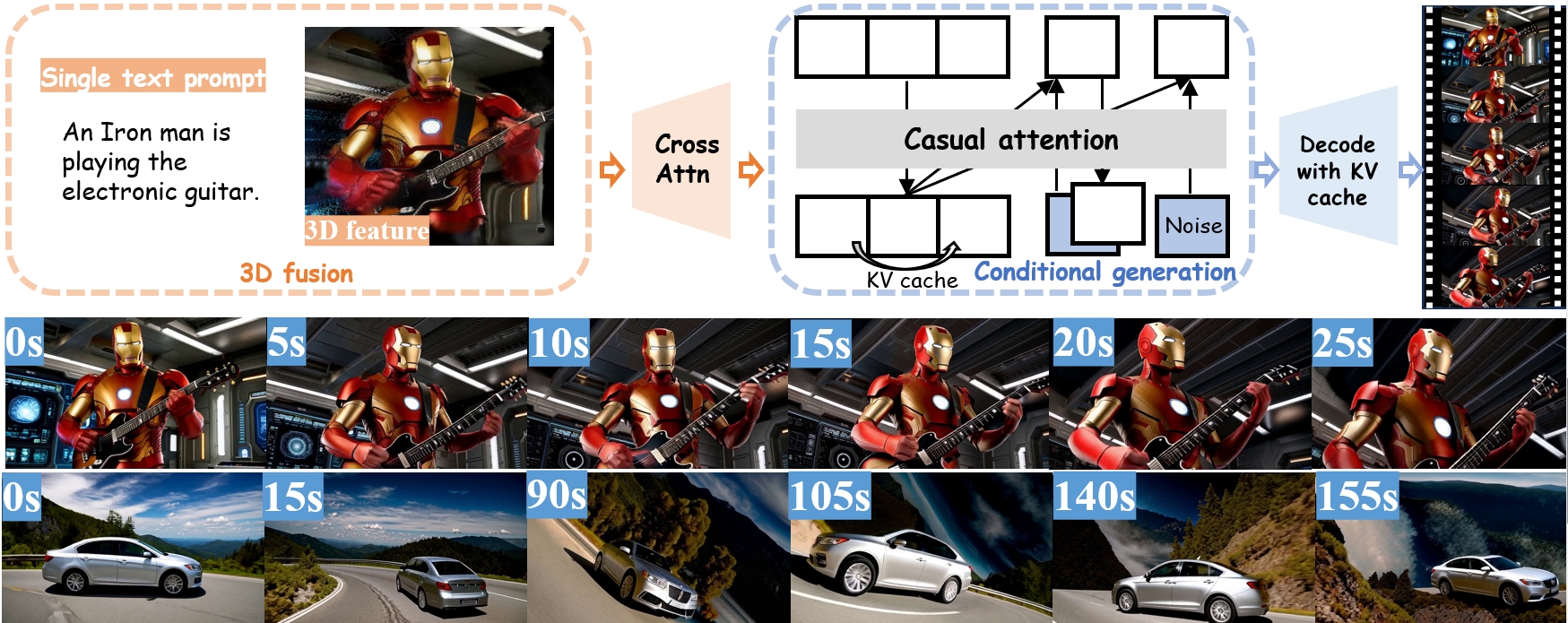}
\captionof{figure}{Endless World features two core modules, 3D fusion and conditional auto-regressive video generation, which together enable the creation of 3D-consistent videos of unlimited length with stable and realistic visual quality.}
\label{fig:teaser}
\vspace{2em}
}]
\begin{abstract}
Producing long, coherent video sequences with stable 3D structure remains a major challenge, particularly in streaming scenarios. Motivated by this, we introduce Endless World, a real-time framework for infinite, 3D-consistent video generation. 
To support infinite video generation, we introduce a conditional autoregressive training strategy that aligns newly generated content with existing video frames. This design preserves long-range dependencies while remaining computationally efficient, enabling real-time inference on a single GPU without additional training overhead.
Moreover, our Endless World integrates global 3D-aware attention to provide continuous geometric guidance across time.
Our 3D injection mechanism enforces physical plausibility and geometric consistency throughout extended sequences, addressing key challenges in long-horizon and dynamic scene synthesis.
Extensive experiments demonstrate that Endless World produces long, stable, and visually coherent videos, achieving competitive or superior performance to existing methods in both visual fidelity and spatial consistency. Our project has been available on \url{https://bwgzk-keke.github.io/EndlessWorld/}.
\end{abstract}    
\section{Introduction}
\label{sec:intro}
Synthesizing long-form temporally coherent videos with consistent 3D structure is a central challenge in computer vision and graphics~\cite{voyager2025,ren2025gen3c}, with broad applications in interactive content creation, virtual reality, and long-horizon simulation~\cite{hunyuan-video}. 
Current video‐diffusion models achieve impressive visual fidelity on short clips \cite{wan,hunyuan-video,cogvideox,mocha,sora,kling,zhang2025thinkdiffuseinfusingphysical,xu2025freevistrainingfreevideostylization,xu2025freevis}, yet they continue to face major quality and consistency issues when extended to long sequences \cite{diffusion-forcing,skyreels-v2,framepack,self-forcing,yang2025longlive}.
Although some long-video generation frameworks (e.g., Self-Forcing~\cite{self-forcing}) use cached key–value states to speed up inference, their quality still degrades over long horizons due to a training–inference discrepancy and the accumulation of small errors.
During training, conditioning frames are trainable, so the model adjusts both new and existing frames. During inference, however, conditioning frames are fixed, leading to errors that accumulate over time.
Furthermore, these approaches typically do not enforce explicit 3D consistency, which can result in artifacts such as unstable geometry, flickering textures, or inconsistent scene layouts across frames. Such issues become particularly pronounced in long‐sequence and infinite‐horizon video generation scenarios.

To address these challenges, we introduce EndlessWorld (Figure~\ref{fig:teaser}), a real-time framework capable of generating infinite, 3D-consistent video. 
First, To address the training-inference discrepancy, we adopt a conditional auto-regressive strategy: new video chunks are generated from existing latent frames with detached gradients, to prevent error propagation.
This design enforces temporal coherence without requiring training on excessively long sequences. Second, we fuse 3D structural features with global text embeddings, allowing 3D information to be globally attended across frames and ensuring consistent geometry and appearance throughout extended sequences. Third, to support infinite-length generation, we introduce an attention sink mechanism that maintains the initial scene characteristics without further fine-tuning. Our Endless World framework achieves real-time inference on a single GPU while preserving high visual fidelity and long-range 3D coherence.
Comprehensive experimental evaluations demonstrate that Endless World generates long, stable, and visually coherent videos, consistently outperforming state-of-the-art methods in terms of both visual quality and 3D consistency.

In summary, our contributions are threefold:
(1) We introduce Endless World, a real-time framework for infinite, 3D-consistent video generation, enabling continuous and stable synthesis over arbitrarily long sequences.
(2) We propose a conditional autoregressive training strategy that enforces temporal and motion consistency without requiring long-sequence training, ensuring efficient and stable generation.
(3) We design a 3D global fusion and scalable attention mechanism that preserves geometric coherence and long-range dependencies, supporting high-quality, real-time video generation on a single GPU.
\section{Related Work}
\label{sec:related_work}

\textbf{Long Video Generation:} 
Recent research has explored hybrid generation frameworks that merge the expressive sampling of diffusion models with the sequential consistency of auto-regressive (AR) predictors~\citep{diffusion-forcing,history-guided,yume,lumos-1,framepack,longvie,streamingt2v,longvie}.
Among these, SkyReels-V2~\citep{skyreels-v2} integrates diffusion-based temporal forcing with a structured film planner and multimodal control interfaces, demonstrating stronger narrative coherence in video synthesis.
Parallel advances in causal AR video generation~\citep{causvid,self-forcing,far,magi-1} emphasize temporally consistent rollout over extended sequences. 
Self-Forcing~\citep{self-forcing} closes the mismatch between training and inference by emulating online roll-out during training, maintaining a persistent key–value cache and conditioning on the model’s own predictions.
Meanwhile, MAGI-1~\citep{magi-1} scales AR video generation to larger model and data regimes via chunk-wise prediction.
Recently, Longlive~\cite{yang2025longlive} enables the use of multiple prompts in an interactive way.  However, its reliance on prompt transitions introduces inefficiencies in long-form generation and temporal flickering.

\noindent\textbf{3D-Consistent Video Generation:} 
Maintaining 3D consistency across video frames is crucial for applications such as interactive content creation, virtual reality, and 3D simulation \cite{team2025hunyuanworld}. A lot of works focus on generating novel views from a set of posed images \cite{nerfstudio,mildenhall2021nerf,kerbl3Dgaussians}, with numerous extensions towards large-scene reconstruction \cite{barron2022mipnerf360,yang2023emernerf,Yu2024GOF,Li_2023_CVPR}, improved rendering quality \cite{adaptiveshells2023,Huang2DGS2024,barron2021mipnerf}, faster rendering speed \cite{adaptiveshells2023,mueller2022instant}, and handling dynamic scenes \cite{luiten2023dynamic,duan:2024:4drotorgs}.
Yet, many of these methods require a dense set of input images and may produce severe artifacts when viewed from extreme viewpoints. Recently, VGGT as an effective method helps to reconstruct high-quality 3D~\cite{wang2025vggt}. 
Existing 3D-aware video generation methods \cite{mallya2020world,ren2025gen3c,voyager2025} often rely on 3D point cloud or 3D cache to synthesize novel views and use it as condition for video synthesis. They are typically limited by the quality of synthesized views. Moreover, most existing methods do not integrate high quality 3D feature guidance directly into the video synthesis process, which is essential for maintaining consistent geometry and appearance during streaming generation.
\section{Method}
\begin{figure*}
    \centering
    \includegraphics[width=\textwidth]{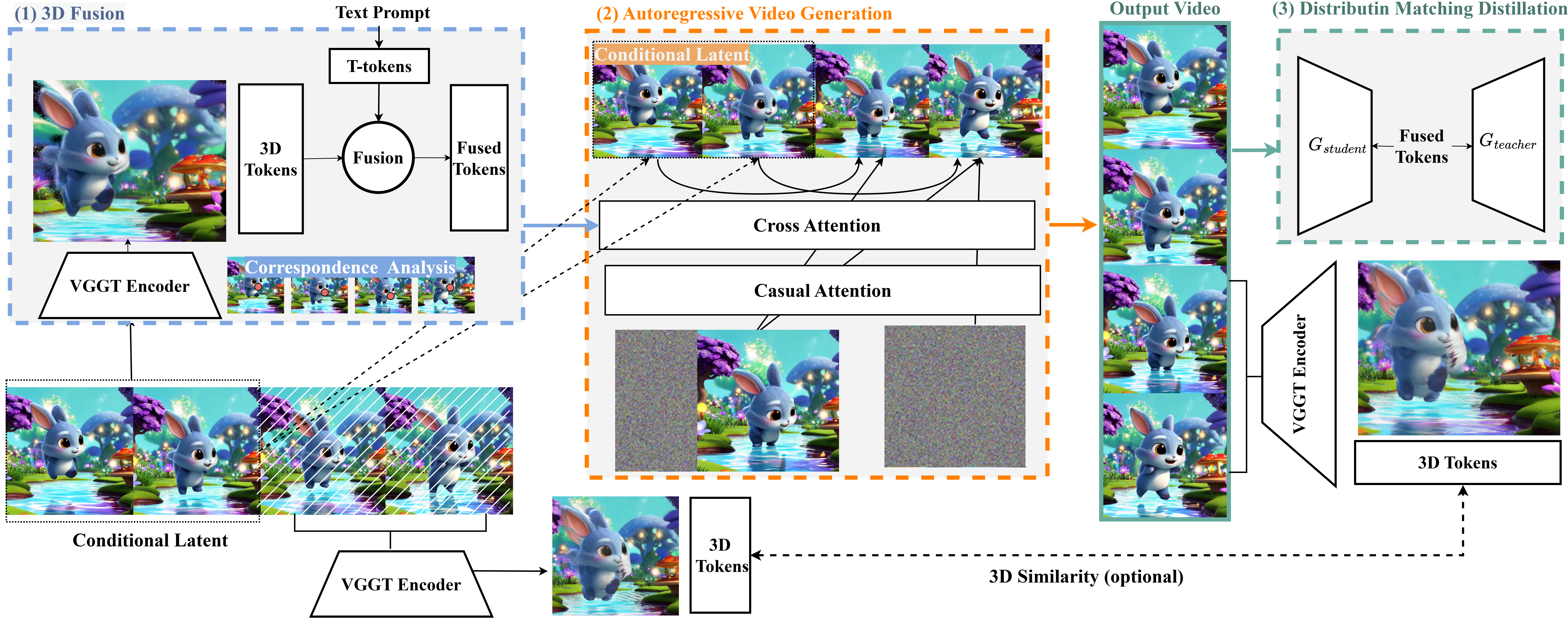}
    \caption{This figure illustrates the pipeline of our proposed Endless World framework. The training process consists of three main stages:
(1) 3D Fusion: fuse 3D features extracted by the VGGT encoder with text tokens;
(2) Conditional Generation: auto-regressively generate video frames conditioned on previously generated frames;
(3) Distribution Matching distillation (DMD): align the distribution of the generated video (including both generated and conditional frames) with that of the supervision video in a training-free manner.
Finally, we optionally apply a 3D similarity loss to ensure consistency between the 3D features of the generated video and those of the reference video with natural motion.}
    \label{fig:pipeline}
\end{figure*}

\label{sec:method}
The core idea of our approach is to leverage 3D structural features to guide video generation while preserving global 3D consistency. To this end, we adopt a conditional generation framework in which each new video frame is synthesized based on previously generated frames (Sec.~\ref{sec:condition}), with gradients detached from prior outputs to prevent error accumulation. We extract 3D features from the previous video content to capture the underlying scene geometry and spatial correspondences. These features are then integrated into the video generation pipeline, providing continuous geometric and structural guidance that promotes coherent and physically plausible video synthesis over extended durations (Sec.~\ref{sec:fuse_3D_cache}). In addition, we introduce an optional 3D consistency loss to balance spatial consistency with overall visual naturalness (Sec.~\ref{sec:optional}). Finally, we employ a streaming generation strategy to support long-horizon video synthesis (Sec.~\ref{sec:streaming}).
Figure~\ref{fig:pipeline} shows the pipeline of our Endless World.

\begin{figure}[!t]
    \centering
    \includegraphics[width=\linewidth]{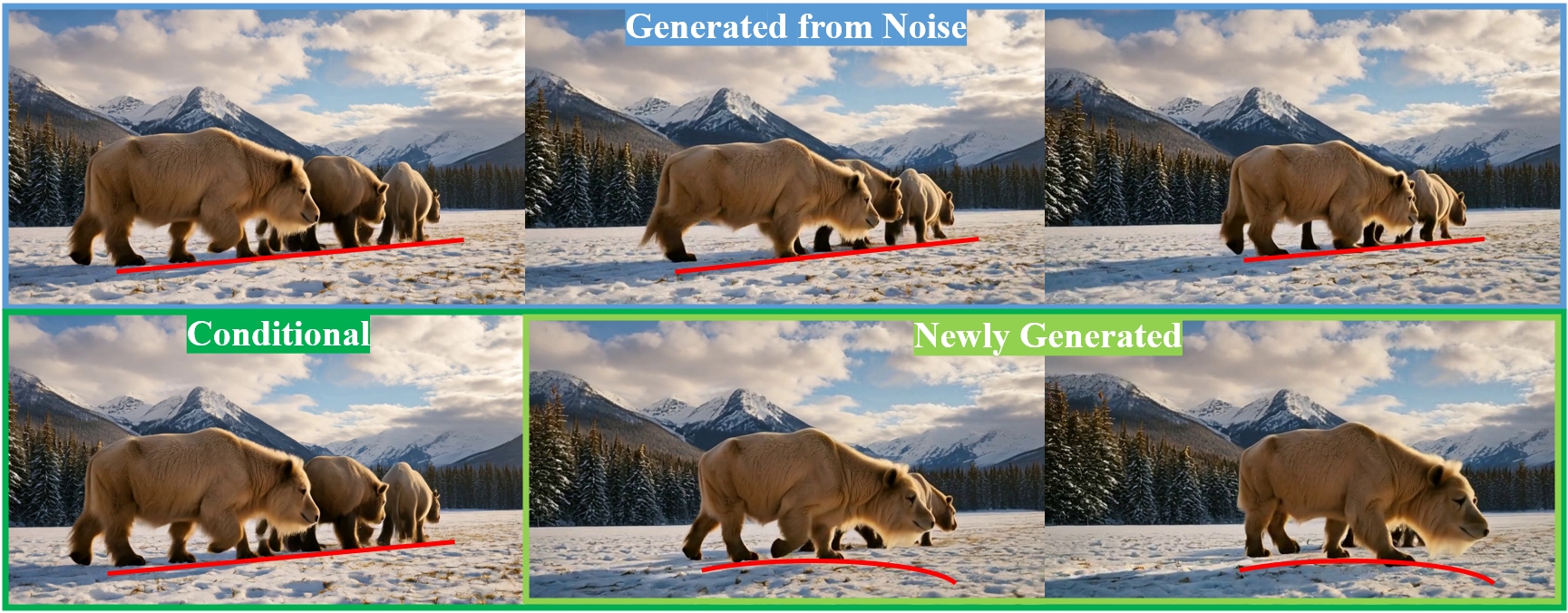}
    \caption{Motion inconsistency in self-forcing autoregressive generation. First row: video generated from noise (cow walks straight). Second row: continuation conditioned on the first video chunk (cow changes direction due to drift).}
    \label{fig:motivation}
\end{figure}

\subsection{Training-Inference Discrepancy}
\label{sec:discrepancy}
Existing self-forcing auto-regressive video diffusion models suffer from a fundamental \textit{training–inference discrepancy}. 
During training, these models are conditioned on previously generated frames and learn to adjust their parameters so that all frames in the sequence appear natural. 
However, because the conditioning frames are differentiable with respect to the model parameters, the gradient updates influence both the conditional and future frames. 
In contrast, during inference, the model generates videos in an auto-regressive manner, where previously generated frames are fixed and cannot be modified. 
This mismatch leads to cumulative temporal errors. As shown in Figure~\ref{fig:motivation}, the first row shows a video generated purely from noise, where the cow walks in a straight line. The second row shows a continuation conditioned on the first video frame, where the cow’s walking direction changes noticeably. Minor deviations in motion accumulate over time, resulting in unnatural drifts, temporal flickering, and physically implausible dynamics.

Let a video sequence be denoted as $v_{1:n} = (v_1, v_2, \dots, v_n)$.  A conventional auto-regressive self-forcing model formulates the joint distribution as
\begin{equation}
    p_\phi(v_{1:n}) = \prod_{k=1}^{n} p_\phi(v_k \mid v^\phi_{<k}),
    \label{eq:ar_chain}
\end{equation}
where \(v^\phi_{<k}\) are the conditioning frames predicted (and differentiable) under the current model parameters \(\phi\). 
Each conditional term \(p_\phi(v_k \mid v^\phi_{<k})\) is modeled by a diffusion-based generator that predicts the next frame conditioned on previous ones.

\begin{figure*}
    \centering
    \includegraphics[width=\textwidth]{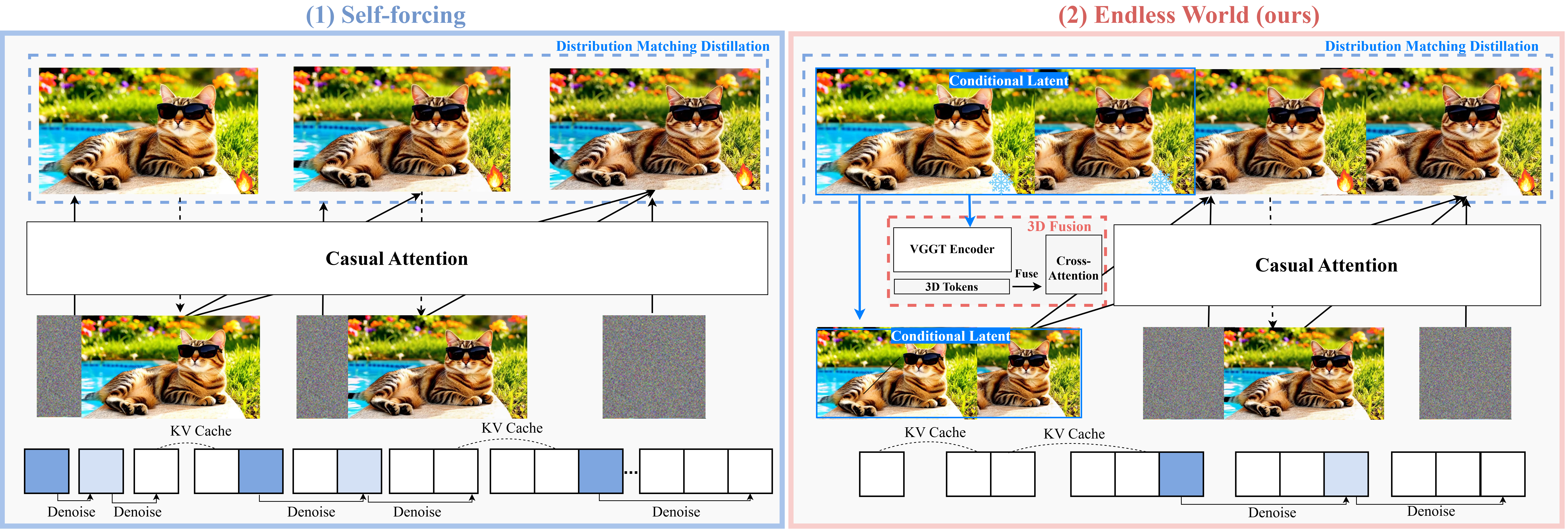}
    \caption{Comparison between Self-Forcing and Endless World.
(1) Self-Forcing autoregressively generates new frames conditioned on previous ones, but jointly optimizes the entire sequence via distribution matching, causing gradients to flow through both past and future frames.
(2) Endless World conditions on existing video frames while restricting gradient updates to newly generated frames, ensuring consistency without altering previous content.}
    \label{fig:vs_forcing}
\end{figure*}

As illustrated in Figure~\ref{fig:vs_forcing}, conventional methods such as Self-forcing~\cite{self-forcing} perform distribution matching distillation (DMD)~\cite{yin2024one, yin2024improved} by first injecting noise into both the generated video distribution \( p_\phi(v_{1:n}) \) and the supervision video distribution \( p_{\text{sup}}(v_{1:n}) \). The distributions are then aligned after applying the forward diffusion process~\cite{jenni2019stabilizing}, denoted as \( p_{\phi, t}(v_{1:n}^t) \) and \( p_{\text{sup}, t}(v_{1:n}^t) \), respectively, where each represents the corresponding distribution at diffusion step \( t \).

We follow~\cite{self-forcing,causvid} using the distribution Matching Distillation (DMD)~\cite{yin2024one, yin2024improved} to minimize the reverse Kullback–Leibler divergence,
by leveraging the score difference between the two distributions to guide parameter updates.
However, since the distribution matching loss matches the entire video and \(v^\phi_{<i}\) depend on \(\phi\), gradients flow through both the conditional frames and the predicted frame. 
This creates a self-forcing effect that alters the conditioning context during optimization, which violates the true inference condition, where \(v_{<i}\) are fixed and non-differentiable.
\subsection{Reformulate Conditional Generation}
\label{sec:condition}
To eliminate this discrepancy, we propose to detach the conditioning frames from the computation graph, treating them as fixed inputs that do not propagate gradients. Figure~\ref{fig:vs_forcing} illustrates the distinction between our approach and Self-forcing.
Formally, we redefine the auto-regressive dependency as:
\begin{equation}
    p_\phi(v_j \mid v^\phi_{i:j-1}, v_{<i}^{\text{detach}}),
    \quad \text{for } j > i,
    \label{eq:cond_general}
\end{equation}
where \(v_{<i}^{\text{detach}}\) indicates that the previous frames are detached from the gradient computation. 
Only the newly generated frames contribute to parameter updates.
The resulting generated sequence becomes:
\begin{equation}
    p_\phi(v_{1:n}) = \prod_{k=i}^{n} p_\phi(v_k \mid v^\phi_{i:k},v_{<i}^{\text{detach}}),
    \label{eq:dmd_loss}
\end{equation}
we then match this sequence with that of supervision data sequence $p_{\text{sup}}(v_{1:n})$.
This formulation ensures that gradient updates align with the auto-regressive inference process,i.e., model parameters are optimized under the same fixed conditioning used at inference time.
By aligning the training condition with the inference behavior, this conditioning strategy prevents gradient leakage across temporal dependencies and stabilizes learning dynamics. 
As a result, the model learns a consistent dynamical prior that maintains smooth motion trajectories and temporal coherence.
\subsection{3D Fusion}
\label{sec:fuse_3D_cache}
For 3D feature fusion, our goal is to use the 3D information extracted by VGGT~\cite{wang2025vggt} as a global representation of the scene, similar with the role of text prompt. Conceptually, both text and 3D features provide high-level guidance about the overall world structure, which can be used to condition video generation. To leverage this, we fuse the global 3D feature into the text embeddings, enabling the video model to incorporate geometric context alongside semantic prompts.

This design achieves two key objectives: (1) the model can be applied with or without 3D global features, eliminating the need to switch models to initialize the scene, and (2) fusing 3D features improves the performance of the video diffusion model by providing additional spatial guidance.

Formally, let $v$ denote the video latent variable and $\hat{f}_{3D} = P_\phi(v)$ be its projected 3D feature. We first compute the 3D feature. 
Let a video latent variable be denoted as $v \in \mathbb{R}^{c\times h \times w \times d}$, where $h$ and $w$ are spatial dimensions and $d$ is the temporal depth of the clip latent, which can be decoded to video sequence of dimension $d'=4(d-1)+1$. For a randomly sampled video clip, we extract 3D features using a pre-trained 3D extractor of VGGT~\cite{wang2025vggt}.  Specifically, for a video $v \in \mathbb{R}^{c \times h \times w \times d'}$, the 3D extractor produces a feature $f_{3D} \in \mathbb{R}^{c' \times h' \times w' \times d'}$, where $c'$ is the 3D feature channel dimension, and $d'$ represents the same temporal dimension. 
We then fuse it with text embedding $e_{\text{text}}$ via a learnable CNN fusion module $f_{\text{fusion}}$:
\begin{equation}
\tilde{e} = f_{\text{fusion}}(e_{\text{text}}, \hat{f}_{3D}),
\end{equation}
where $\tilde{e}$ denotes the fused embedding used to condition video generation. In the learnable fusion module, we first project the 3D features through a convolution layer to match the dimensionality of the text embedding. The resulting features are then passed through a zero convolution layer and added to the text embedding to form the fused representation. This fusion module is jointly optimized with the distribution-matching distillation loss to enhance the overall quality of the generated videos.
This setup allows the model to incorporate 3D structural guidance without retraining the generation backbone, resulting in more coherent and geometrically consistent videos.

\subsection{3D Similarity}
\label{sec:optional}
We introduce a soft 3D coherence constraint based on a cosine similarity loss, rather than explicitly aggregating multi-view information as in prior work.
During training, we perform two-step generation: (1) generate the full video sequence, and (2) randomly mask a subset of frames, predicting them from the previous conditional context.
Let $\hat{v}_t$ denote a predicted frame conditioned on context and $v_t$ the corresponding frame generated purely from noise. Both are projected into the 3D feature space via $P_\theta$, yielding $\hat{f}_{3D}^t = P_\theta(\hat{v}_t)$ and $f_{3D}^t = P_\theta(v_t)$.
The 3D consistency loss is defined as:
\begin{equation}
\mathcal{L}_{\text{3D}} = 1 - \frac{\langle \hat{f}_{3D}^t, f_{3D}^t \rangle}{|\hat{f}_{3D}^t|_2 , |f_{3D}^t|_2},
\label{eq:3D_reg}
\end{equation}
where $\langle \cdot , \cdot \rangle$ denotes the dot product. This encourages alignment between predicted and reference 3D representations, promoting temporal and geometric coherence without enforcing strict reconstruction.
While this loss improves 3D structural consistency across frames and viewpoints, it may slightly reduce temporal smoothness and perceptual naturalness. Therefore, we employ it as an optional module, enabling a controllable balance between geometric fidelity and visual realism.

\noindent \textbf{Streaming and Long-Horizon Video Generation.} 
\label{sec:streaming}
Motivated by streaming LLM~\cite{streamingllm}, we apply an attention sink to preserve contextual memory during inference time. 
We also apply the rotational positional embedding after the key–value (KV) cache. This strategy encodes temporal phase continuity between consecutive generation windows, ensuring smooth transitions and mitigating boundary artifacts in continuous video synthesis.

\section{Experiment}
\subsection{Setup}

\textbf{Model Training:}
The model is trained using 3D structural features from VGGT~\cite{wang2025vggt}, extracted from videos decoded from latent variables initialized with random noise.
Given a video sequence consisting of $81$ frames, we divide it into temporal blocks of three frames each and apply random temporal masking to simulate conditional prediction. 
Specifically, we mask a future segment $\{t, \dots, T\}$, where $T = 81$ and $t < T$ is divisible by $3$, 
and use the preceding unmasked frames as the conditioning context for predicting the masked portion.
During training, the 3D features extracted from the unmasked frames are projected and fused with the corresponding text embeddings to enhance geometric-semantic alignment. 
We optionally apply a 3D consistency regularization term $\mathcal{L}_{\text{3D}}$, as defined in Eq.~\ref{eq:3D_reg}, to encourage spatial coherence across frames. 
The overall training objective is formulated as:
\begin{equation}
\mathcal{L}_{\text{total}} = \mathcal{L}_{\text{gen}} + \lambda_{\text{3D}} \mathcal{L}_{\text{3D}},
\end{equation}
where $\lambda_{\text{gen}}$ indicates the generation loss calculated by distribution matching distillation. $\lambda_{\text{3D}}$ denotes the weight parameter that controls the contribution of the 3D regularization term.  We empirically set $\lambda_{\text{3D}}=0.1$. 

\textbf{Model Inference:}
During inference, we initialize the video latent with Gaussian noise and generate subsequent frames in an auto-regressive manner. 
We fuse text embeddings with 3D features extracted by VGGT from the video to provide 3D structural guidance. The resulting fused embeddings are then used for video extension, without adding any extra computational overhead.
For long video sequences, we employ an attention sink mechanism that preserves the tokens of the first frame, ensuring temporal continuity throughout the generation process.
To enable efficient long-horizon synthesis, we alternate between two conditioning modes:
(1) Long-context conditioning: we condition on 18 latent representations (including one attention-sink frame and 68 recent frames, i.e., $(18 - 1) \times 4 = 68$) to generate 3 latent variables (12 frames).
(2) Short-context conditioning: we condition on 3 latent representations (the attention-sink frame and the two most recent latents, equivalent to the first frame and 8 latest frames) to generate 18 new latent steps (72 frames).
This alternating strategy balances temporal coherence with computational efficiency, enabling continuous generation of long, temporally consistent videos.

\textbf{Dataset and Evaluation metrics:}
To comprehensively evaluate both visual fidelity and semantic consistency, we adopt VBench-long~\cite{zheng2025vbench2} and conduct a user preference study (details provided in the supplementary material).
For the VBench evaluation, we generate 944 videos following the standard VBench prompt pipeline used in prior works~\cite{self-forcing,yang2025longlive}, and assess our model across 16 evaluation dimensions covering both video quality and semantic alignment using VBench-long.
Specifically, the video quality metrics include temporal flickering, background consistency, subject consistency, dynamic degree, imaging quality, aesthetic quality, and motion smoothness.
The semantic consistency metrics cover color, human action, multiple objects, object class, overall consistency, temporal style, appearance style, scene, and spatial relationship.

\begin{table*}[t]
  \caption{
Baseline comparison on VBench.
Endless World is compared with publicly available video generation models of similar scale and resolution.
Scores are reported on the standard VBench~\cite{huang2023vbench} prompt set, with baseline results taken from~\cite{yang2025longlive}. The FPS is evaluated with a single H100 GPU.
  }
  \label{tab:short}
  \centering
\resizebox{\textwidth}{!}{
\begin{tabular}{lcccccccc}
  \toprule
  \multirow{2}{*}{Model} & \multirow{2}{*}{Video Length}&\multirow{2}{*}{Type}&\multirow{2}{*}{Params} & \multirow{2}{*}{Resolution} & \multirow{2}{*}{\makecell{Throughput\\(FPS) $\uparrow$}} & \multicolumn{3}{c}{Evaluation scores $\uparrow$}\\
  \cmidrule(lr){7-9}
   &  &  &  & & & Total & Quality & Semantic \\
  \midrule
  LTX-Video~\citep{HaCohen2024LTXVideo} &5s & Diffusion & 1.9B & $768{\times}512$ & 8.98          & 80.00 & 82.30 & 70.79 \\
  Wan2.1~\citep{wan}        &5s & Diffusion & 1.3B & $832{\times}480$ & 0.78          & 84.26 & 85.30 & 80.09 \\
  MAGI-1~\citep{magi-1}     &5s     & Autoregressive            & 4.5B & $832{\times}480$ & 0.19          & 79.18 & 82.04 & 67.74 \\
  CausVid~\citep{causvid}   &5s    & Autoregressive             & 1.3B & $832{\times}480$ & 17.0 & 81.20 & 84.05 & 69.80 \\
  NOVA~\citep{deng2024nova}  &5s     & Autoregressive          & 0.6B & $768{\times}480$ & 0.88          & 80.12 & 80.39 & 79.05 \\
  Pyramid Flow~\citep{pyramid-flow}  &5s & Autoregressive       & 2B   & $640{\times}384$ & 6.7           & 81.72 & 84.74 & 69.62 \\
    \midrule
  SkyReels-V2~\cite{skyreels-v2} &30s& Autoregressive&1.3B & $960{\times}540$ &0.49 & 75.29 & 80.77 & 53.37 \\
 FramePack~\cite{framepack}  &30s & Autoregressive& 1.3B & $960{\times}540$ & 0.92& 81.95 & 83.61  & 75.32 \\
 Self-Forcing~\cite{self-forcing} &30s& Autoregressive& 1.3B & $832{\times}480$ & 17.0 & 81.59 & 83.82 & 72.70 \\
 LongLive~\cite{yang2025longlive}   &30s  & Autoregressive& 1.3B & $832{\times}480$ & 20.7 &\underline{83.52} & \underline{85.44} & \underline{75.82} \\
\midrule
  \multirow{1}{*}{\textbf{Endless World}}&30s& Autoregressive& 1.3B & $832{\times}480$ & 17.0 &\textbf{84.54} & \textbf{85.52} & \textbf{80.60} \\
  \bottomrule
\end{tabular}}
\label{tab:comparison}
\end{table*}
\textbf{Implementation details:}
We develop Endless World upon the Wan2.1-T2V-1.3B backbone~\cite{wan}, a text-to-video model capable of generating 5-second clips at 16 FPS with a resolution of 832 × 480. To enable efficient autoregressive generation, we convert the pretrained model into a few-step causal-attention framework following the self-forcing DMD paradigm ~\cite{self-forcing}, trained on the VidProM dataset~\cite{wangvidprom}. During inference, we activate the frame sink mechanism by retaining all tokens from the initial frame as persistent sink tokens. Additionally, we utilize VGGT~\cite{wang2025vggt} to extract high-quality 3D structural features for spatial consistency. We use four NVIDIA H100 GPUs for model training and one NVIDIA H100 GPU for model inference. 

\subsection{Long Video generation comparison}
We begin our evaluation by testing Endless World on VBench using its official long-video generation prompts. For fair comparison, we benchmark against a suite of open-source video generation models of comparable scale, including LTXVideo~\cite{HaCohen2024LTXVideo}, Wan 2.1~\cite{wan}, SkyReels-V2~\cite{skyreels-v2}, MAGI-1~\cite{magi-1}, CausVid~\cite{causvid}, NOVA~\cite{deng2024nova}, PyramidFlow~\cite{pyramid-flow}, Self-Forcing~\cite{self-forcing}, and LongLive~\cite{yang2025longlive}.
All evaluations follow VBench’s standardized and normalized scoring protocol to ensure consistency and comparability across models.

\textbf{Quantitative comparison (30 seconds):} On 30-second clips, Endless World achieves performance exceeding the strongest baselines, demonstrating high visual fidelity and temporal stability (Table~\ref{tab:comparison}). Note that several methods, such as LTXVideo, Wan 2.1, MAGI-1, CausVid, NOVA, and PyramidFlow, are constrained to 5-second generations; for these, we report their 5-second results for reference. For auto-regressive models capable of longer video synthesis, we include 30-second results for SkyReels-V2, FramePack, Self-Forcing, and Longlive, following comparative settings from~\cite{yang2025longlive}. Across all metrics, Endless World surpasses existing methods by a significant margin in both visual quality and semantic alignment, confirming that our model sustains consistent generation quality over extended durations.

\textbf{Quantitative comparison (60 seconds):}
For further evaluate long-term generation capability, we assess Endless World on the 60-second VBench benchmark and compare it against recent autoregressive baselines. 
As shown in Table~\ref{tab:compare60vs30}, Endless World preserves high visual quality even for 60-second generations, surpassing the self-forcing model’s 30-second results across all evaluation metrics.
For broader comparison, we additionally include SkyReels-V2, Self-Forcing, and LongLive. 
The results of these methods are taken directly from~\cite{yang2025longlive}, which reports VBench performance on 160 videos with interactive prompts. 
Since Longlive does not specify the exact prompt set used in these evaluations, we provide our own results on 944 videos from the official VBench dataset as an intermediate reference for comparison. 
Notably, prior approaches exhibit significant quality degradation as video duration increases, highlighting their limitations in maintaining temporal consistency and geometric stability. 
For completeness, we also report results for Self-Forcing variants incorporating attention sink. 
As summarized in Table~\ref{tab:quality_metrics}, Endless World achieves the highest overall performance across all quality metrics, demonstrating superior stability, visual fidelity, and temporal coherence over long-horizon video generation.

\begin{figure*}
    \centering
    \includegraphics[width=\textwidth]{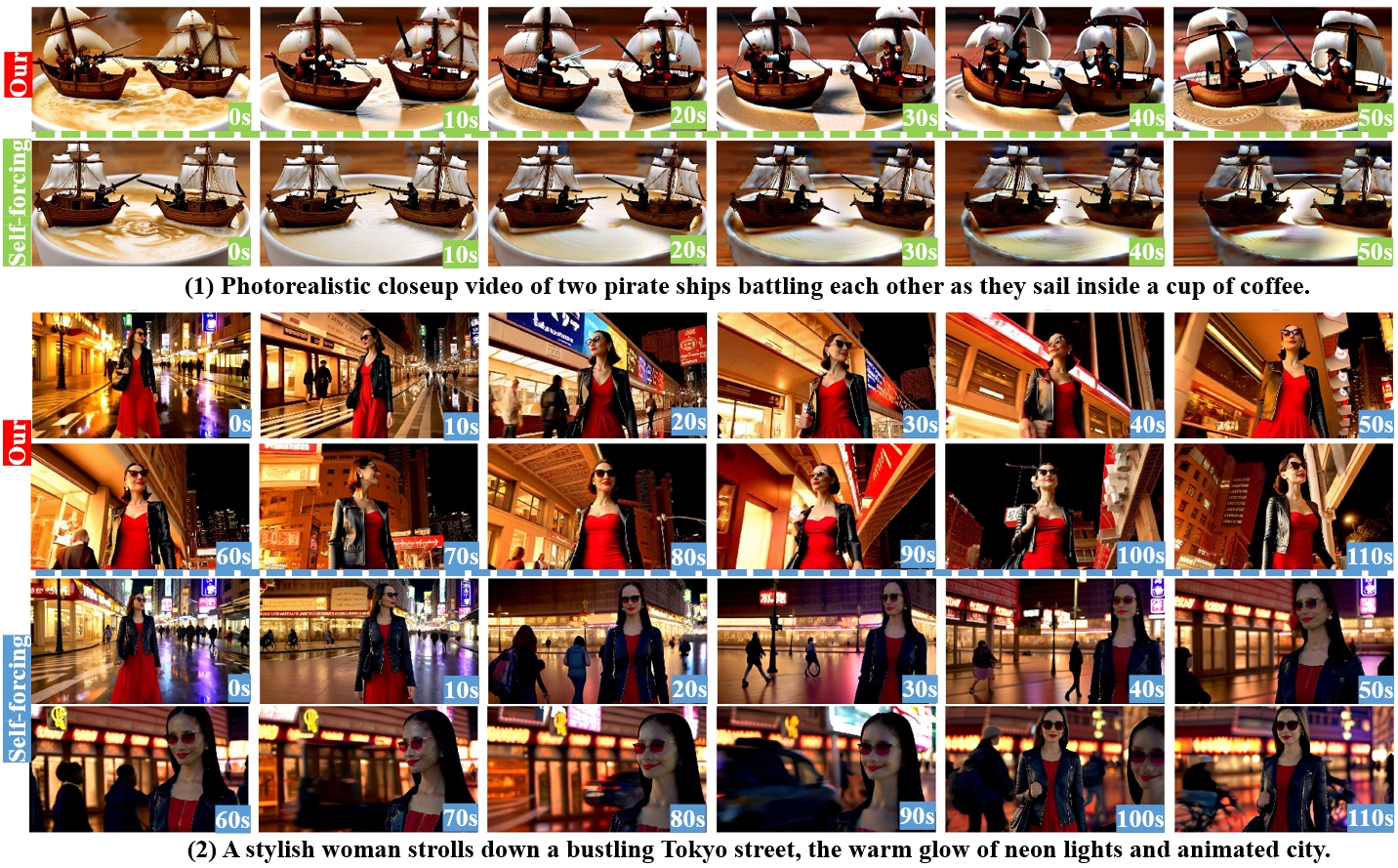}
   \vskip-10pt \caption{Comparison of long-duration video generation.
We compare Endless World with Self-Forcing (with attention sink) for one- and two-minute sequences.
Endless World preserves visual quality and temporal coherence throughout, whereas Self-Forcing suffers from progressive quality degradation.}
    \label{fig:visual}
\end{figure*}

\textbf{Qualitative Comparison (120 seconds):}  
Figure~\ref{fig:visual} presents a qualitative comparison between our method and the Self-Forcing baseline equipped with attention sink.  
We compare both 60-second and 120-second generated videos.  
While Self-Forcing exhibits severe degradation in visual quality beyond approximately 20 seconds, manifested as color shifts, structural distortions, and flickering artifacts, our approach preserves consistent color tone, style, and motion dynamics throughout the entire sequence.  
These qualitative results further validate the robustness and long-term stability of the proposed Endless World framework. 


\begin{table}[t]
\centering
\caption{Comparison of 60-second video generation on VBench.
Using interactive results from~\cite{yang2025longlive} as reference, we compare Self-Forcing and our Endless World on single-prompt generation.}
\resizebox{\linewidth}{!}{
\begin{tabular}{lcccc}
\toprule
Method &Video Length &Interactive & Sink  & Quality Score $\uparrow$ \\
\midrule
SkyReels-V2 & 60s&$\checkmark$ & $\times$ & 80.49 \\
Self-forcing& 60s&$\checkmark$ & $\times$ & 82.46 \\
LongLive & 60s&$\checkmark$ & $\checkmark$ & 84.38 \\
\midrule
Self-forcing& 30s&$\times$ & $\times$& 83.82\\
Self-forcing& 30s&$\times$ & $\checkmark$& 83.89\\
Endless World &60s&$\times$ & $\checkmark$&\textbf{84.73}\\
\bottomrule
\end{tabular}}
\label{tab:quality_metrics}
\end{table}

\begin{table}[t]
\centering
\caption{Ablation study of key components on 30-second video generation.
We evaluate the impact of attention sink, conditional generation, and 3D fusion using the VBench benchmark.}
\resizebox{\linewidth}{!}{
\begin{tabular}{lcccccc}
\toprule
\multirow{2}{*}{Sink} &\multirow{2}{*}{Condition} & \multicolumn{2}{c}{3D} & \multicolumn{3}{c}{Evaluation Score}\\
\cmidrule(lr){3-4}\cmidrule(lr){5-7}
&&Video Latent&Text Token &Total $\uparrow$ & Quality $\uparrow$ & Semantic $\uparrow$ \\
\midrule
$\times$&$\times$& $\times$&$\times$& 81.59& 83.82 & 72.70\\
$\checkmark$&$\times$& $\times$&$\times$& 82.94& 83.89 & 79.16\\
$\checkmark$&$\checkmark$& $\times$&$\times$& 83.30 & 84.50 & 78.48 \\
$\checkmark$&$\checkmark$& $\checkmark$&$\times$& 82.83 & 84.10 & 77.73 \\
\midrule
$\checkmark$&$\checkmark$& $\times$&$\checkmark$& \textbf{84.54} & \textbf{85.52} & \textbf{80.60} \\
\bottomrule
\end{tabular}}
\label{tab:ablation}
\end{table}

\subsection{Ablation studies}
We perform a series of ablation experiments to investigate the contributions of key design components, including attention sink, conditional generation, 3D fusion strategy, 3D similarity loss, and video length.\\
\noindent\textbf{Effect of Attention Sink.}  
As shown in Table~\ref{tab:ablation}, integrating the attention sink significantly enhances spatiotemporal coherence, increasing the overall VBench score from 81.59 to 82.94. In particular, the semantic dimension shows a notable improvement, with the semantic score rising from 72.70 to 79.16. This indicates that stabilizing long-range attention is critical for preserving semantic consistency and mitigating temporal quality drift during extended video generation.



\textbf{Effect of Conditional Generation.}  
We first evaluate the contribution of conditional generation to the overall model performance.  
As reported in Table~\ref{tab:ablation}, enabling conditional generation yields a significant improvement in total score from 82.94 to 83.30.
This enhancement suggests that conditioning each prediction on preceding frames encourages better temporal alignment and preserves fine spatial details over long sequences.  
In particular, the model becomes more robust to cumulative drift, producing visually stable and coherent results throughout extended video durations.  

\textbf{Impact of 3D Fusion.}  
Next, we analyze the contribution of the 3D consistency fusion module.
Integrating this component yields additional performance gains, raising the overall VBench score to 84.54. As shown in Table~\ref{tab:3D_detail}, exploiting 3D structural cues helps the model maintain stable spatial representations over time, reducing issues such as object jitter or distortion. This leads to clear improvements in spatial and object consistency, as well as aesthetic quality.
Taken together, these results demonstrate that conditional generation and 3D fusion jointly play a critical role in producing high-quality videos with strong temporal stability and geometric coherence.
\begin{table}[!t]
\centering
\caption{Effect of incorporating 3D fusion on VBench. “Objects” denotes multi-objects, and “Spatial” measures spatial relationship.}
\resizebox{\linewidth}{!}{
\begin{tabular}{lccccc}
\toprule
3D Fusion & Aesthetic Quality  &  Objects & Spatial & Overall Consistency \\
\midrule
$\times$ & 61.72  & 81.73 & 78.59 & 71.60 \\
$\checkmark$ & \textbf{66.33} & \textbf{90.55} & \textbf{79.84} & \textbf{74.42} \\
\bottomrule
\end{tabular}}
\label{tab:3D_detail}
\end{table}

\textbf{Fusion Strategy: Text Embedding vs. Video Latent:} 
We also compare two fusion strategies, \textit{i.e.}, injecting the 3D features into video latent variables or into text embeddings. Fusion at the global text level produces more stable and higher-quality results, while latent-level fusion, despite preserving 3D geometry, disrupts local motion and introduces flow inconsistencies and flickering. 

\begin{table}[!t]
\centering
\caption{Effect of the 3D similarity loss on 30-second VBench.}
\resizebox{\linewidth}{!}{
\begin{tabular}{lcccccccc}
\toprule
\multirow{2}{*}{$\mathcal{L}_{3D}$}&\multicolumn{3}{c}{Consistency ($\uparrow$)}&\multicolumn{2}{c}{Smoothness ($\uparrow$)}\\
\cmidrule(lr){2-4}\cmidrule(lr){5-7}
 &Subject & Background & Temporal & Motion & Aesthetic Quality \\
\midrule
$\times$&93.89 & 94.79 & 97.86 & \textbf{95.05} & \textbf{66.33} \\
$\checkmark$& \textbf{96.32} & \textbf{94.95} & \textbf{98.41} & 94.84 & 61.60 \\
\bottomrule
\end{tabular}}
\label{tab:regularization}
\end{table}

\textbf{Effect of 3D similarity loss:} As presented in Table~\ref{tab:regularization}, incorporating the 3D regularization loss enhances the consistency-related metrics, including subject consistency, background consistency, and temporal flickering. Nevertheless, this improvement comes at the expense of metrics associated with perceptual smoothness, such as motion smoothness and aesthetic quality. This observation suggests that while the 3D regularization effectively enforces geometric coherence across frames, it simultaneously introduces a trade-off between geometric fidelity and perceptual naturalness, leading to slightly less smooth and visually appealing video outputs.

\textbf{Detailed Dimension Analysis.}  
To gain deeper insight into the contributions of each design component, we conduct a detailed dimension-wise evaluation covering both quality and semantic aspects using VBench.  
The quality dimensions include subject consistency, background consistency, temporal flickering, motion smoothness, aesthetic quality, imaging quality, and dynamic degree.  
The semantic dimensions encompass object class, multiple objects, human action, color, spatial relationships, scene, appearance style, temporal style, and overall consistency.  

\begin{table}[t]
\centering
\caption{Impact of video length on Vbench Scores.}
\resizebox{\linewidth}{!}{
\begin{tabular}{lcccc}
\toprule
\multirow{2}{*}{Method} & \multirow{2}{*}{Video Length} &\multicolumn{3}{c}{Evaluation Score}\\
\cmidrule(lr){3-5}
&&Total $\uparrow$ & Quality $\uparrow$ & Semantic $\uparrow$ \\
\midrule
\multirow{2}{*}{Self-Forcing} & 5s & 84.31& 85.07 & 81.28\\
& 30s & 81.59& 83.82 & 72.70\\
\hdashline
\multirow{2}{*}{Endless World}& 30s & \textbf{84.54}& \textbf{85.52} & \textbf{80.80}\\
& 60s & 82.31& 84.73 & 72.63\\
\bottomrule
\end{tabular}}
\label{tab:compare60vs30}
\end{table}
\begin{figure}[!t]
  \centering
  \begin{minipage}[t]{0.49\linewidth}
    \centering
    \includegraphics[width=\linewidth]{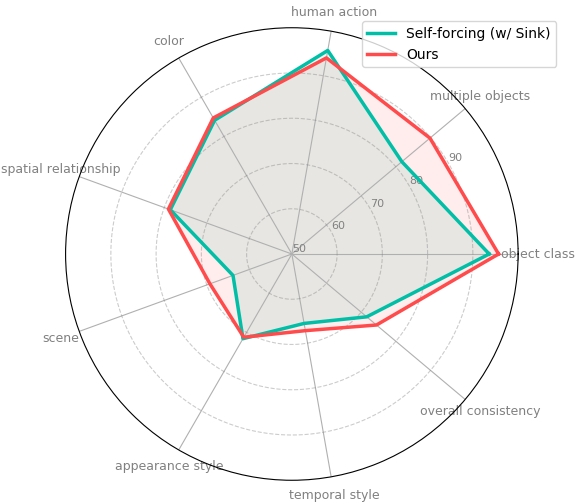}
    \caption{Semantic ablations.}
    \label{fig:modelA}
  \end{minipage}
  \hfill
  \begin{minipage}[t]{0.49\linewidth}
    \centering
    \includegraphics[width=\linewidth]{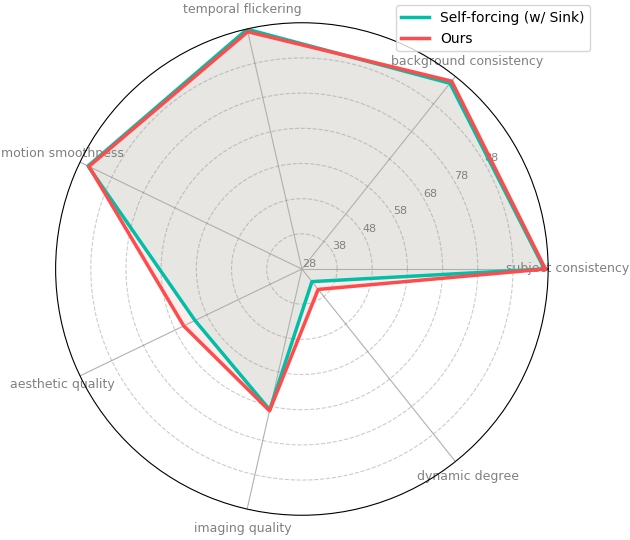}
    \caption{Quality ablations.}
    \label{fig:modelB}
  \end{minipage}
\end{figure}
As shown in Figure~\ref{fig:modelA} and Figure~\ref{fig:modelB}, incorporating 3D features leads to consistent improvements across nearly all dimensions.  
Notably, the 3D fusion module enhances semantic alignment while simultaneously strengthening visual fidelity, yielding more coherent, natural, and balanced video generation results.  

\textbf{Effect of Video Length.}  
We further analyze the robustness of our model with respect to generation duration.  
Table~\ref{tab:compare60vs30} reports results for videos of different lengths.  
Unlike prior methods, which suffer from significant performance degradation as video length increases, our model maintains stable quality across extended sequences.  
This indicates that the proposed conditional autoregressive training and 3D fusion mechanisms effectively prevent cumulative temporal drift, enabling reliable long-horizon video synthesis.  

\section{Discussion}
In this work, we presented Endless World, a framework for real-time, geometry-aware autoregressive video generation.
The proposed conditional autoregressive mechanism provides a scalable and efficient solution for streaming video synthesis, enabling adaptability to open-world and interactive scenarios.
Our 3D fusion strategy enables global 3D-aware attention, effectively preserving long-term structural stability and mitigating temporal drift.
Extensive experiments on public benchmarks demonstrate that Endless World achieves significant improvements over existing methods in both visual quality and object consistency.
{
    \small
    \bibliographystyle{ieeenat_fullname}
    \bibliography{main}
}


\end{document}